A high-precision self-supervised monocular visual odometry in foggy weather based on robust cycled generative adversarial networks and multi-task learning aided depth estimation

Xiuyuan Li[※], Jiangang Yu, Fengchao Li, Guowen An
School of Instrument and Electronics, North University of China, Taiyuan, China
E-mail: lixiuyuannuca@126.com

Abstract:
　　This paper proposes a high-precision self-supervised monocular VO, which is specifically designed for navigation in foggy weather. A cycled generative adversarial network is designed to obtain high-quality self-supervised loss via forcing the forward and backward half-cycle to output consistent estimation. Moreover, gradient-based loss and perceptual loss are introduced to eliminate the interference of complex photometric change on self-supervised loss in foggy weather. To solve the ill-posed problem of depth estimation, a self-supervised multi-task learning aided depth estimation module is designed based on the strong correlation between the depth estimation and transmission map calculation of hazy images in foggy weather. The experimental results on the synthetic foggy KITTI dataset show that the proposed self-supervised monocular VO performs better in depth and pose estimation than other state-of-the-art monocular VO in the literature, indicating the designed method is more suitable for foggy weather.

1 Introduction

　　High-precision self-contained navigation is in great demand for Unmanned Aerial Vehicles (UAV) and Unmanned Ground Vehicles (UGV) which always operate in environments where GNSS is unreliable or unavailable [1, 2]. Visual odometry (VO) is an effective independent navigation device in GNSS-denied environments because it is not easily interfered with and screened by electromagnetic waves [3, 4]. Due to the smaller size, lower hardware cost and less calibration effort compared with binocular VO, monocular VO has drawn significant attentions in industry and academia during the last few years, showing promising results in terrestrial and aerial vehicles in mild weather [5-7].

　　However, navigation via monocular VO in foggy weather is more challenging because the captured images are hazy due to light attenuation and scattering, resulting in difficulty in acquiring effective feature points [8-9]. Consequently, traditional feature-based monocular VO is prone to fail because the keypoints of hazy images are hard to track and match correctly, even if stable and robust feature detectors/descriptors are employed, such as the scale invariant feature transform (SIFT), the speeded-up robust features (SURF) or feature descriptors based on the Zernike moments. Besides, direct monocular VO is not robust enough because direct methods estimate motion via minimizing the photometric error, which is significantly influenced by the uneven distribution of fog and brightness in foggy weather.

Due to the excellent capability in feature extraction, deep learning based monocular VO developed in recent years [10-12] have already shown promising performance in terms of both accuracy and robustness in challenging environments, where traditional monocular VO cannot work well. In consideration that ground truth poses are difficult and expensive to obtain, many self-supervised deep learning monocular VO have been proposed to explore the possibility of learning without full supervision [13-15]. A common assumption adopted by most self-supervised learning methods is photometric consistency, that is, the photometric error of corresponding pixel of the same object in adjacent frames is zero [16]. Unfortunately, the photometric consistency assumption is often violated by brightness change resulted from poor imaging conditions in foggy weather. Thus, the location accuracy of most existing self-supervised monocular VO designed for urban scenes in mild weather may decrease greatly if they are directly used on foggy scenarios due to the photometric inconsistency caused by the inhomogeneous and changing brightness in foggy imaging conditions.

To alleviate the adverse influence of photometric inconsistency, some effective methods have been proposed in recent years. GeoNet [17] added structural similarity (SSIM) to loss to mitigate the effects of brightness change. SSIM captures more local information than the sum of absolute photometric differences, but it does not capture global information. D3VO [18] directly predicted the affine parameters through a network to globally adjust the brightness of the warped images adaptively. However, D3VO cannot deal with the local brightness change, which often occur due to the sharply uneven illumination. On the other hand, this kind of transformation based methods cannot exactly remove the brightness discrepancies which are normally nonlinear and hard to be modelled in the real scene. There also exists some novel explorations, such as transferring original frames into a new feature space which is unaffected by brightness and season changes [19]. However, the feature space is not exactly photometric consistent for complicated scenarios, such as foggy environments.

Besides, in self-supervised monocular VO, estimation of depth and pose are simultaneously learned in a coupled way, thus, accurate depth contributes to precise pose estimation and vice versa. However, depth estimation from a monocular image is essentially an ill-posed problem because a monocular image alone does not contain sufficient physical cues for scene depth [20]. Thus, the depth estimation of most previous self-supervised monocular VO is vague, resulting that the estimation accuracy is worse than that of supervision methods [21].

In order to improve the accuracy and robustness of monocular VO for foggy scenarios, a self-supervised deep monocular VO based on robust cycled generative adversarial networks and multi-task learning aided depth estimation is proposed in this paper according to the characteristics of hazy images in foggy weather. In contrast to previous self-supervised monocular VO based on generative adversarial networks (GAN) in a straightforward way, a cycled generative network structure is designed in this paper to further improve the equality of self-supervised signal. The proposed cycled frame conducts the image reconstruction from different views in a closed loop, implicitly constraining on each other. In the presented frame, not only the different-view reconstruction loss helps for better optimization of the pose estimation network, but also the two estimated poses are connected with a consistence loss to provide strong supervision from each half cycle.

To eliminate the interference of complex photometric change on self-supervised loss in foggy weather, gradient-based loss and perceptual loss are introduced to handle the local and global illumination change, respectively. Gradient-based loss is generated from the photometric differences between the gradient-based feature maps of the warped and target frames, on account of the robustness of gradient-based representation to local illumination changes. Perceptual loss is calculated from the VGG-feature distance between warped and target frames, based on the empirical observation that the classification results by VGG models are not very sensitive when the pixel intensity range is manipulated, which is concurred by another recent study [22]. Consequently, perceptual loss constrains the feature information of the images except illumination, resulting in the robustness to global illumination change.

Inspired by the strong correlation between the depth estimation and transmission map calculation of hazy images in foggy weather, a multi-task learning aided depth estimation module in a self-supervised way is presented in this paper, aiming at enjoying the mutual benefits between these two related tasks to obviously promote the accuracy of depth estimation. The self-supervised loss for transmission map calculation is generated from the differences between the raw and corresponding synthetized hazy images, which are reconstructed based on the imaging model in foggy weather. As a consequence, the precision of pose estimation can be significantly improved resulting from much more accurate depth estimation.

The experimental results demonstrate that the proposed self-supervised deep-learning monocular VO can significantly improve the estimation accuracy of depth and pose in foggy weather compared with previous works in the literature. The presented self-supervised framework based on cycled GAN can help monocular VO to obtain much more accurate self-supervised signal than the common GAN in a straightforward way. The introduced gradient-based loss and perceptual loss can offer self-regularization with respect to illumination changes, resulting in more robust self-supervised signal to local and global brightness changes between adjacent images. The proposed multi-task learning aided depth estimation module can significantly improve the accuracy of depth estimation via the joint learning of related transmission map calculation, leading to more accurate pose estimation.

## 2 Related Works

Monocular VO has been considered as a multi-view geometric problem for decades. It is traditionally solved by minimizing photometric or geometric reprojection errors and works well in regular environments, but the reliability and robustness of such schemes in challenging scenarios (e.g. textureless scenes or abrupt motions) still face great difficulties. In view of these limitations, monocular VO has been studied with deep learning methods in the last few years and many approaches with promising performance have been proposed.

A. Supervised monocular VO

To achieve accurate estimation of the camera poses, supervised monocular VO treat it as a supervised learning problem and many methods with good results have been proposed. DeMoN [23] treated the structure from motion (SfM) as a learning problem, and jointly estimated the depth and camera poses in an end-to-end supervised fashion. Zhou et al. [24] used a coarse-to-fine deep learning framework to achieve full 6-DOF keyframe pose tracking and dense mapping. To alleviate drift errors which accumulated over time, many state-of-the-art approaches adopted recurrent neural network (RNN) to use temporal information. Wang et al. presented DeepVO [25] to learn the camera poses from image sequences via combining CNNs and LSTM RNNs. MagicVO [12] calculated camera poses based on a sequence of continuous monocular images via a hybrid CNN and bi-directional LSTM framework.

The limitation of supervised monocular VO is that a large amount of labeled data are necessary for the training. The acquisition of ground truth always requires expensive equipment or highly manual labeling, and some ground truth is even unable to obtain, such as optical flow. Data synthesis is proposed to solve the problem of insufficient ground truth, but there exists always gap between synthetic and real-world data, limiting the accuracy of the supervised monocular VO trained with synthetic data.

B. Self-supervised monocular VO

In order to alleviate the reliance on ground truth, recently many self-supervised methods have been proposed for monocular VO. Zhou et al. [16] presented the first self-supervised monocular VO SfMLearner, which leverages the geometric correlation of depth and pose to learn both of them in a coupled way. Inherited from this idea, many self-supervised monocular VO have been proposed, including modifications on loss functions [26, 27], network architectures [13, 26, 28, 29, 30], predicted contents [17], and combination with classic VO/SLAM [31, 32]. Recently, Almalioglu et al. [33] combined a recurrent learning approach with GAN for pose and depth estimation in a self-supervised manner. Similarly, Li et al. [34] adopted GAN to generate image-level loss to significantly improve the equality of the self-supervised loss.

Despite its feasibility, self-supervised monocular VO still underperforms supervised ones mainly due to the weak robustness of the self-supervised loss and the strongly ill-posed characteristic of coupled depth estimation. Researchers have proposed some outstanding methods, which can obviously improve the robustness of the self-supervised loss for the training of monocular VO designed for urban scenes in mild weather [17-19]. However, these methods play a limited role in foggy weather, because they do not fully consider the specific global and local photometric inconsistency between adjacent frames due to the inhomogeneous and changing brightness in foggy imaging conditions. In addition, the accuracy of depth estimation is associated with the accuracy of pose estimation in the existing coupled self-supervised framework. The inaccuracy of depth estimation caused by the strongly ill-posed characteristic inevitably lead to the low accuracy of pose estimation. Unfortunately, the solution of the ill posed problem of self-supervised monocular depth estimation is still an open topic.

## 3 Proposed methods

### A. The overview of the proposed self-supervised monocular VO in foggy weather

The framework of self-supervised monocular VO in foggy weather in this paper is shown in Fig.1. The generator of the forward half-cycle produces the synthesized previous image based on the real previous image and current image. In the backward half-cycle, the generator takes as inputs the synthesized previous image and real current image, predicting the synthesized current image. For the two synthesized images, the corresponding discriminators are adopted to identify if the images are fake or true. The discriminators try to distinguish the synthesized images from the real ones. The generator finishes the training until the discriminators cannot distinguish the synthesized images from the real ones. The consistence loss is added between the two estimated 6-DoF poses generated separately from the forward or backward half-cycle, putting a strong constraint for each half-cycle and thus facilitating the learning of both half-cycles.

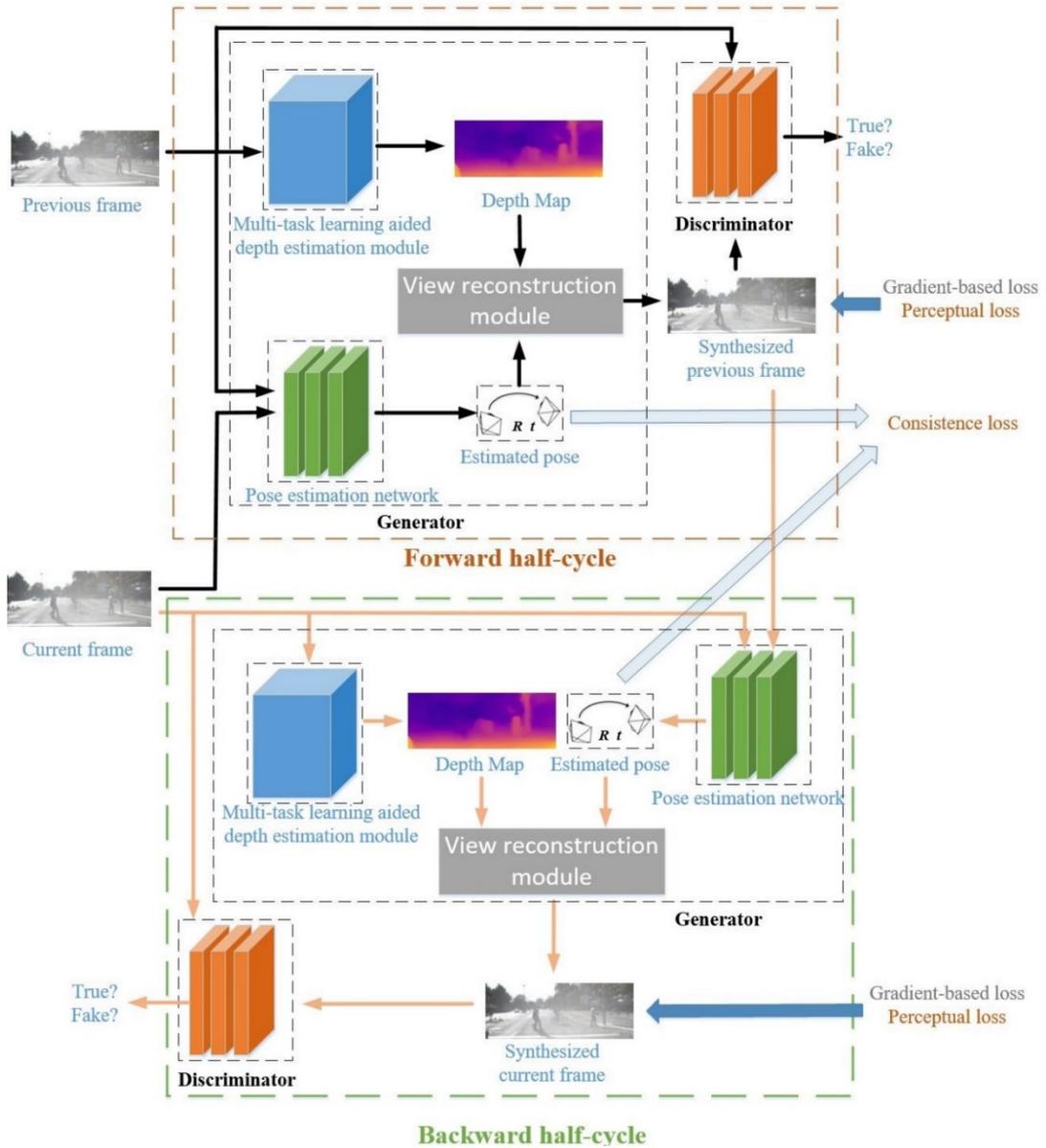

Figure 1. The proposed framework of self-supervised monocular VO in foggy weather

The generator consists of the proposed multi-task learning aided depth estimation module, pose estimation network and view reconstruction module. The pose estimation network includes two parts: CNN and LSTM, where the CNN determines the most discriminative visual features from the image batch generated by stacking the two adjacent underwater images, and then the LSTM takes in input the feature vector and estimated previous 6-DoF pose to predict the current 6-DoF pose. In parallel, the depth estimation module generates a depth map of the target view. The view reconstruction module synthesizes the target image using the generated depth map, estimated 6-DoF pose and nearby pixel values from the source image.

B. Pose estimation network

To learn effective features from a set of two consecutive monocular images for the pose estimation, the CNN-based feature extractor is built on the ResNet18 architecture [35] with the following modifications. The structure is similar to the ResNet18 truncated before the last average pooling layer. Each of the convolutions is followed by batch normalization and Rectified Linear Unit (ReLU). The architecture replaces the average pooling with global average pooling and subsequently adds two inner-product layers which output a visual feature vector representation.

In order to model temporal dependencies to derive accurate 6-DoF pose regression, a two-layer Bi-directional LSTM with 1000 hidden states is used as the LSTM, which takes in input the combined feature representation at current time step and its previous hidden states. After the LSTM network, a fully connected layer serves as the 6-DoF pose regressor, mapping the features to a pose transformation, representing the motion transformation over a time window.

C. View reconstruction module

The target image can be reconstructed based on the estimated depth map of the target image and estimated 6-DoF pose between the two adjacent images. Suppose $p_2$ denotes a pixel in the target image that is also visible in the source image, its projection $p_1$ on the source image is represented by

$$p_1 \sim K_1 \left[ \hat{R}_{12} \middle| \hat{t}_{12} \right] \hat{D}_2(p_2) K_2^{-1} p_2 \qquad (1)$$

where ~ means 'equal in the homogeneous coordinate'. $K_1$ and $K_2$ are the intrinsic matrix for the corresponding two images, respectively. $\left[ \hat{R}_{12} \middle| \hat{t}_{12} \right]$ is the estimated 6-DoF pose between the two adjacent images via the pose estimation network. $\hat{D}_2(p_2)$ is the estimated depth map of the target image via the multi-task learning aided depth estimation module.

Thus, in the view reconstruction module, the synthesized image can be obtained using the source image by bilinear sampling according to Eq.(1), based on the outputs of the pose estimation network and multi-task learning aided depth estimation module.

D. Principle of the proposed self-supervised multi-task learning aided depth estimation module

In 1976, Mccartney proposed an atmospheric scattering model [36] to describe foggy images according to the scattering effect of atmospheric particles on light. And then Narasimhan developed the model and presented the widely used hazy imaging model in foggy weather as follow [37]:

$$I(x) = J(x)t(x) + A(1-t(x))$$
$$t(x) = e^{-\beta d(x)} \quad (2)$$

where $x$ is a pixel, $I(x)$ is the real hazy image in foggy weather, $J(x)$ is the clear image, $A$ is the background light, and $t(x)$ is the transmission map which represents light attenuation due to the scattering medium. The level of attenuation is determined by the attenuation coefficient $\beta$ and scene depth $d(x)$.

As can be seen from Eq.(2), the depth map is closely related to the transmission map of hazy images in foggy weather, indicating that the tasks of transmission map calculation and depth estimation from a single hazy image have a strong connection essentially.

Besides, the training dataset with enough number of real paired images in foggy weather for transmission map calculation is not easy to acquire because it is daunting even impossible to collect a large scale desirable ground truth in foggy environments in practice. Although the methods based on synthetic images can alleviate the problem of lack of ground truth, the synthesized transmission maps are probably inconsistent with the real transmission maps, which would lead to the domain shift issue when the model trained on the synthetic dataset is applied to the real-world hazy images in foggy weather. Thus, it is highly expected to develop a novel transmission map calculation network which could work in a self-supervised way, while achieving the promising performance.

Inspired by the strong correlation between the depth map and transmission maps of hazy images in foggy weather, a self-supervised joint multi-task framework for depth estimation and transmission map calculation is proposed based on the hazy image formation model in foggy weather in Eq.(2), aiming at enjoying the mutual benefits between these two related tasks to improve the estimation accuracy of depth maps in a self-supervised way. As depicted in Fig.2 , the raw input image is firstly processed by an encoder, which is a fine-tuned version of the ResNet50 network pruned of its fully connected layers and outputs the shared features. Features are then fed to two task-dependent decoders: a depth map estimation decoder and a transmission map calculation decoder. The two decoders share the same architecture with different weights to accomplish the corresponding estimation. Each of the two decoders uses skip connections from the encoder's activation blocks, enabling it to resolve higher resolution details. The detailed structure of the adopted encoder and decoders in this paper is inspired by and similar to the encoder and decoder of the depth estimation network in [38], respectively.

In parallel, the raw hazy image is input to a separate encoder-decoder, which has the same architecture to the encoder-decoders for depth map estimation and transmission map calculation but different weights, to estimate the corresponding clear image. As the background light is independent of the image content and owns the global property, it is estimated directly in this paper by the method [39], which is based on robust statistical model and can achieve high-precision estimation, thus reducing the training difficulty of the whole network.

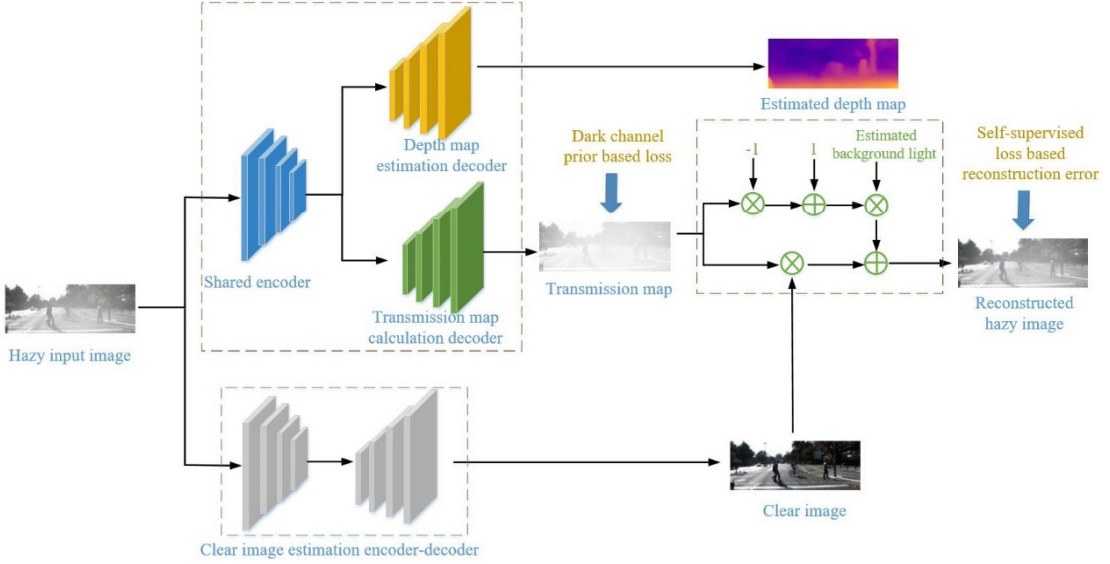

Figure 2. The proposed self-supervised multi-task learning aided depth estimation module

According to the hazy image formation model in Eq.(2), the real hazy image in foggy weather can be reconstructed based on the estimated transmission maps, clear image and background light. Consequently, the self-supervised reconstruction loss for transmission map estimation is generated from the differences between the original and reconstructed hazy images, and can be expressed as:

$$L_{\text{Re}c} = \left\| I_{\text{Re}c}(x) - I_{ori}(x) \right\| \qquad (3)$$

where $\left\| \cdot \right\|$, $I_{ori}(x)$ and $I_{\text{Re}c}(x)$ denote Frobenius norm of a given matrix, original image and reconstructed image, respectively.

The reconstruction loss can only constrain the reconstructed hazy image to be well close to the real one, but cannot ensure that the outputs of the transmission map calculation decoder are always consistent with the real transmission maps. Thus, an auxiliary loss is introduced in this paper to further constrain the transmission map calculation decoder. The auxiliary loss is defined as the differences between the calculated transmission map by the decoder and the approximate estimation of transmission maps via traditional model-based methods, and can be written as:

$$L_{aux} = \left\| t(x)_{dec} - t(x)_{tra} \right\| \qquad (4)$$

where $t(x)_{dec}$ and $t(x)_{tra}$ denote the calculated transmission map by the decoder and the approximate estimation of transmission maps via traditional model-based methods [39], respectively.

E. Self-supervised losses of robust cycled GAN

In addition to conventional pixel-level losses, the proposed self-supervised losses in this paper also include the image-level losses generated via both half-cycles of the cycled GAN, unique cycled consistence loss of pose estimation, gradient-based loss and perceptual loss, to promote the accuracy and robustness of the trained monocular VO in foggy weather.

To ensure the pixel-level correspondence between the synthesized and original image, a pixel-level reconstruction loss is established from both weighted photometric loss $L_{pho}$ and structural similarity metric (SSIM) [40] as follows:

$$L_p = L_{reg}(\hat{M}) + (1-\alpha) L_{pho} + \frac{1}{N} \sum_{x,y} \alpha \frac{SSIM(\hat{I}(x,y), I(x,y))}{2} \quad (5)$$

Where $L_{reg}(\hat{M})$ is a regularization term to prevent the network from converging to a trivial solution, which is detailed in [16]. $N$ is the number of images in the training minibatch. $\hat{I}(x,y)$ and $I(x,y)$ are the intensity value of the pixel $(x,y)$ in the synthesized and original image, respectively.

Besides, a pixel-level edge-aware smoothness loss is introduced to enforce discontinuity and local smoothness in depth considering that discontinuity of depth usually happens where strong image gradients are present. Similar to [29], the edge-aware smoothness loss can be expressed as:

$$L_s = \frac{1}{N} \sum_{x,y} (\|\nabla_x \hat{D}(x,y)\| e^{-\|\nabla_x I(x,y)\|} + \|\nabla_y \hat{D}(x,y)\| e^{-\|\nabla_y I(x,y)\|}) \quad (6)$$

Where $\hat{D}(x,y)$ is the estimated depth map of the pixel $(x,y)$.

Different to the GAN in a straightforward way, the image-level adversarial losses for the cycled GAN in this paper contain two parts: the loss calculated from the discriminator of the forward half-cycle and the loss generated from the discriminator of the backward half-cycle, leading to much more accurate and robust self-supervised signal. The discriminator is a convolutional network, whose architecture is the same as that of the discriminator adopted in [33]. Importantly, the discriminators with the same functionality for the two half-cycles share their parameters in order to avoid an increase in the number of parameters. In order to stabilize the training, this paper adopts a least-squares GAN loss [41] by substituting the cross-entropy loss via the least-squares function with binary coding (1 for real, 0 for synthesized). Thus, the adversarial loss for the forward half-cycle can be formulated as follows:

$$L_{gan}^{D_{gan},f}(D_{gan}) = E_{I_{pre} \sim p(I_{pre})} \left[ (D_{gan}(I_{pre}) - 1)^2 \right]$$
$$+ E_{I_{cur}, \hat{D}_{pre} \sim p(I_{cur}, \hat{D}_{pre})} \left[ (D_{gan}(G_{gan}(I_{cur}, \hat{D}_{pre})))^2 \right] \quad (7)$$

$$L_{gan}^{G_{gan},f}(D_{gan}) = E_{I_{cur}, \hat{D}_{pre} \sim p(I_{cur}, \hat{D}_{pre})} \left[ (D_{gan}(G_{gan}(I_{cur}, \hat{D}_{pre})) - 1)^2 \right] \quad (8)$$

Where $L_{gan}^{D_{gan},f}$ and $L_{gan}^{G_{gan},f}$ are the losses for the training of the discriminator $D_{gan}$ and generator $G_{gan}$. $I_{pre}$ and $I_{cur}$ are the real previous and current image, respectively. $\hat{D}_{pre}$ is the estimated depth map of the real previous image.

The Eq.(7) represents that the discriminator is trained to output 1 when the input image is a real image and 0 when the input is a synthesized image. In Eq.(8), the generator is trained in order to predict an image such that the discriminator confuses the synthesized images with real images, thus outputting 1.

The adversarial loss for the backward half-cycle is acquired similarly and the total adversarial loss for the cycled GAN can be expressed as:

$$L_{gan} = L_{gan}^{G_{gan},f} + L_{gan}^{D_{gan},f} + L_{gan}^{G_{gan},b} + L_{gan}^{D_{gan},b} \quad (9)$$

Where $L_{gan}^{D_{gan},b}$ and $L_{gan}^{G_{gan},b}$ are the losses for the discriminator $D_{gan}$ and generator $G_{gan}$ in the training of the backward half-cycle, respectively.

In order to make the two 6-DoF pose estimations generated by the forward and backward half-cycle constrain on each other, a unique cycled consistence loss of pose estimation is proposed and formulated as follows:

$$L_{cyc} = \left\| \hat{p}_f - \hat{p}_b \right\| \quad (10)$$

Where $\hat{p}_f$ and $\hat{p}_b$ are the estimated 6-DoF poses via the forward and backward half-cycle, respectively.

Compared with most of terrestrial and aerial environments in mild weather, the illumination between adjacent frames in foggy imaging conditions often changes more dramatically due to the inhomogeneous and changing brightness in foggy weather, decreasing the robustness of the above designed self-supervised losses based on the assumption of photometric inconsistency. Thus, some self-supervised losses that can effectively eliminate the interference of illumination changes are necessary for the self-supervised training of robust cycled GAN.

On one hand, a self-supervised loss based on gradient feature representation is designed for the training to deal with local lighting changes between adjacent frames. The gradient magnitudes are used, instead of raw intensities, in direct VO and have proven to be robust with the local lighting changes [42], which always happen in foggy weather. Inspired by [43], the gradient-based loss can be calculated from the gradient magnitude differences between the two consecutive frames and expressed as follows:

$$L_{gra} = \left\| \nabla \hat{I}(x, y) - \nabla I(x, y) \right\| \tag{11}$$

Where $\nabla \hat{I}(x, y)$ and $\nabla I(x, y)$ denote the gradient vector of the synthesized image $\hat{I}(x, y)$ and real image $I(x, y)$ from the image intensities around a given pixel $(x, y)$, respectively.

On the other hand, in order to alleviate the influence of global illumination changes on training losses, a self-supervised perceptual loss is introduced on account of the experimental results that VGG models can preserve the image content features to itself though the light changes to some extent, which is observed in [22]. Thus, the perceptual loss is defined as the VGG-feature distance between real and synthesized images, constraining the feature information of the images except illumination, and can be expressed as:

$$L_{per} = \frac{1}{W_{i,j} H_{i,j}} \sum_{x=1}^{W_{i,j}} \sum_{y=1}^{H_{i,j}} (\phi_{i,j}(\hat{I}) - \phi_{i,j}(I))^2 \tag{12}$$

where $\hat{I}$ and $I$ are the synthesized and real image, respectively. $\phi_{i,j}$ denotes the feature map extracted from a VGG-16 model pre-trained on ImageNet. $i$ represents the $i$-th max pooling of the VGG-16 model, and $j$ represents the $j$-th convolutional layer after the $i$-th max pooling layer. $W_{i,j}$ and $H_{i,j}$ are the dimensions of the extracted feature maps. By default, the $i$ and $j$ are set as 5 and 1, respectively.

The final loss consists on the combination of the pixel-level losses, image-level losses via both half-cycles, cycled consistence loss of pose estimation, gradient-based loss and perceptual loss, and can be written as follows:

$$L_{final} = \lambda_p L_p + \lambda_s L_s + \lambda_{gan} L_{gan} + \lambda_{cyc} L_{cyc} + \lambda_{gra} L_{gra} + \lambda_{per} L_{per} \tag{13}$$

where $\lambda_p, \lambda_s, \lambda_{gan}, \lambda_{cyc}, \lambda_{gra}$ and $\lambda_{per}$ are the corresponding weights for controlling the importance of different losses.

4  Experimental Results and Discussions

A. Datasets

Considering that the training of deep networks requires a large number of samples and it is not easy to obtain enough real foggy samples containing a variety of sensor data for navigation research, a synthetic foggy dataset is created based on the publicly-available dataset KITTI [44] to test the performance of the proposed improvements and method. The KITTI dataset includes a full set of input sources including raw images, 3D point cloud data from LIDAR and camera trajectories. This dataset contains 11 sequences collected from a vehicle driving around a residential area with ground-truth poses and depth available (and 11 sequences without ground-truth).

The corresponding synthetic foggy image of each image in the KITTI dataset was obtained based on the raw clear image and depth map according to the method in [45]. The parameters used in the synthesis process in this paper are the same as those in [45]. After the foggy images corresponding to each image in the KITTI dataset are acquired, a foggy KITTI dataset containing 11 sequences with multiple sensor information and ground truth is established, which is suitable for the evaluation of monocular VO in foggy weather.

B. Training Details

The neural nets were implemented using the Pytorch framework and trained on a Nvidia 3090 Ti GPU. The sequences 00-08 of the foggy KITTI dataset was used for training, while the sequences 09 and 10 were adopted for testing. All the training and testing images were from the left monocular camera of the stereo pair. The whole network is trained end-to-end by the Adam algorithm with an initial learning rate of 0.0001. During the training, dropout of 0.3 was applied to avoid over-fitting.

C. Design of ablation and comparative experiments

In order to verify the improvement effects of the proposed multi-task learning aided depth estimation module and network architecture based on robust cycled GAN, some variations of the proposed self-supervised monocular VO in foggy weather were designed via removing the corresponding parts in the ablation study. The different methods and variations designed for the ablation study are listed as below:

Baseline: The baseline for the ablation study is a simplified version of the proposed complete method by modifications in two aspects. On one hand, the overall architecture is simplified via replacing the robust cycled GAN by GAN in a straightforward way without the proposed cycled consistence loss of pose estimation, gradient-based loss and perceptual loss. That is to say, the losses for the training of the baseline only consist of the pixel-level losses and image-level losses via the forward half-cycle. On the other hand, the depth estimation network of the baseline is just a common encoder-decoder, which only contains one decoder for the depth estimation, thus losing the promotion from transmission map calculation to the depth estimation.

Baseline with cycled GAN: Based on the baseline, this variation is obtained via updating the GAN in a straightforward way to a cycled GAN. Thus, the losses for the training of this variation contain the pixel-level losses, image-level losses via both half-cycles and cycled consistence loss of pose estimation;

Baseline with robust cycled GAN: Based on the baseline with cycled GAN, this version is acquired by adding gradient-based loss and perceptual loss for the training to alleviate the interference of illumination changes between adjacent frames in foggy weather;

Baseline with multi-task learning aided depth estimation module: Based on the baseline, this variation is established via substituting the simple depth estimation network with the proposed multi-task learning aided depth estimation module;

Proposed complete method: This version is the proposed complete method in this paper. Compared with the baseline, the proposed complete method has three improvements: cycled GAN based architecture bringing the image-level losses via both half-cycles and cycled consistence loss of pose estimation into the training, gradient-based loss and perceptual loss which promote the robustness of losses to illumination changes, and multi-task learning aided depth estimation module.

Besides, some representative existing monocular VO in the literature, namely, SfMLearner [16], Zhan et al. [13], Vid2Dpeth [30] and GeoNet [17] were selected and tested on the foggy KITTI dataset, to further validate the performance of the proposed complete method, which is specifically designed for foggy weather.

D. Depth Estimation Results

The estimation accuracy of depth has direct effect on the equality of self-supervised signal, further affecting the results of pose estimation. Thus, some variations of the proposed method were tested to verify the improvement effects of the designed cycled GAN based architecture, robust losses against brightness changes, and multi-task learning aided depth estimation module.

As illustrated in Tab.1, the baseline with multi-task learning aided depth estimation module shows obviously better results than the baseline, indicating the effectiveness of multi-task learning aided depth estimation module. In the multi-task learning framework, the coupled transmission map calculation decoder adds constraints to the shared encoder and depth estimation decoder, which makes the depth estimation more accurate. The qualitative examples of depth estimation by different variations are shown in Fig.3, in which most variations have difficulty in recovering the depth of small objects and mistake the depth of several objects. As illustrated in Fig.3, the estimated depth map by the baseline with multi-task learning aided depth estimation module can preserve more accurate details than the baseline, such as boundaries and thin structures, due to the constraints from the designed multi-task learning framework.

Table 1. The depth estimation results of ablation experiments
(The results are capped at 50m. As for error metrics Abs Rel, Seq Rel, RMSE and RMSE log, lower value is better; as for accuracy metrics $\delta < 1.25$, $\delta < 1.25^2$ and $\delta < 1.25^3$, higher value is better.)

| Method | Cap | Abs Rel | Sq Rel | RMSE | RMSE log | $\delta < 1.25$ | $\delta < 1.25^2$ | $\delta < 1.25^3$ |
|---|---|---|---|---|---|---|---|---|
| Baseline | 50m | 0.205 | 1.392 | 6.545 | 0.310 | 0.559 | 0.627 | 0.648 |
| Baseline with cycled GAN | 50m | 0.187 | 1.279 | 5.891 | 0.283 | 0.575 | 0.639 | 0.661 |
| Baseline with robust cycled GAN | 50m | 0.174 | 1.185 | 5.565 | 0.265 | 0.594 | 0.652 | 0.667 |
| Baseline with multi-task learning aided depth estimation module | 50m | 0.168 | 1.126 | 5.472 | 0.257 | 0.601 | 0.659 | 0.672 |
| Proposed complete method | 50m | 0.136 | 0.933 | 4.647 | 0.232 | 0.635 | 0.667 | 0.682 |

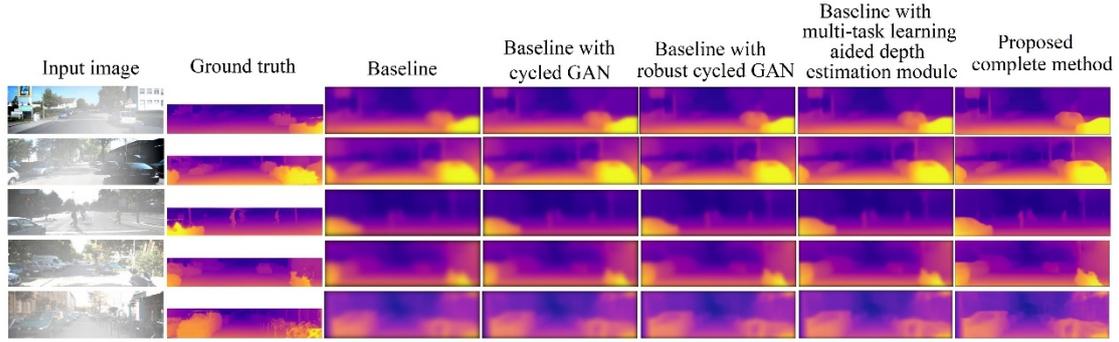

Figure 3. The qualitative examples of depth estimation in ablation experiments

Compared with the baseline based the pixel-level losses and image-level losses via the forward half-cycle, the baseline with cycled GAN can remain more edges and details in the depth map and obtain more accurate depth estimation because the cycled GAN has more constraints and stronger consistency than the GAN in a straightforward way does. The baseline with cycled GAN obtains worse results than the baseline with robust cycled GAN, illustrating that the adopted robust losses against brightness changes are beneficial to depth estimation to some extent.

Besides, the performance of the proposed complete method was evaluated compared with some state-of-the-art deep learning based monocular depth estimation methods. The quantitative and qualitative results of depth estimation by different methods are shown in Tab.2 and Fig.4 respectively, which indicates the proposed complete method outperforms all other approaches in foggy weather.

Table 2. The depth estimation results by different methods

| Method | Cap | Abs Rel | Sq Rel | RMSE | RMSE log | $\delta < 1.25$ | $\delta < 1.25^2$ | $\delta < 1.25^3$ |
|---|---|---|---|---|---|---|---|---|
| SfMLearner [16] | 50m | 0.397 | 2.226 | 7.823 | 0.369 | 0.467 | 0.611 | 0.652 |
| Vid2Dpeth [30] | 50m | 0.227 | 1.413 | 6.575 | 0.325 | 0.544 | 0.623 | 0.647 |
| GeoNet [17] | 50m | 0.213 | 1.407 | 6.552 | 0.321 | 0.549 | 0.620 | 0.642 |
| Zhan et al. [13] | 50m | 0.210 | 1.398 | 6.551 | 0.315 | 0.555 | 0.618 | 0.637 |
| Proposed complete method | 50m | 0.136 | 0.933 | 4.647 | 0.232 | 0.635 | 0.667 | 0.682 |

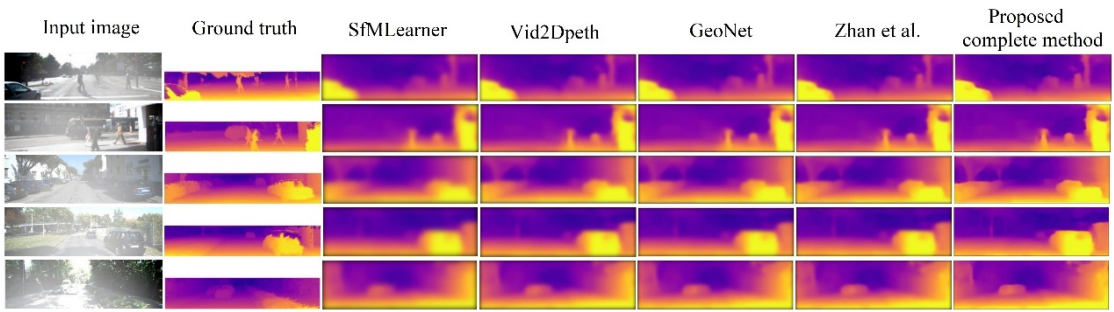

Figure 4. The qualitative examples of depth estimation by different methods

As illustrated in the above qualitative and quantitative comparisons, the proposed complete method achieves better performance than other state-of-the-art methods owing to the cycled GAN based architecture, robust losses against brightness changes, and multi-task learning framework, which demonstrates the effectiveness of the designed model to infer the scene depth from a single hazy image in foggy weather.

E. Pose Estimation Results

To analyze the influence of the cycled GAN based architecture, robust losses against brightness changes and multi-task learning framework on the pose estimation accuracy, some variations of the proposed method were applied to the dataset and the corresponding results are shown in Tab. 3.

Table 3. The pose estimation results of ablation experiments

| Method | Absolute trajectory error on foggy KITTI sequence 09 (m) | Absolute trajectory error on foggy KITTI sequence 10 (m) |
| --- | --- | --- |
| Baseline | $0.025 \pm 0.015$ | $0.028 \pm 0.016$ |
| Baseline with cycled GAN | $0.021 \pm 0.010$ | $0.023 \pm 0.015$ |
| Baseline with robust cycled GAN | $0.018 \pm 0.0085$ | $0.019 \pm 0.010$ |
| Baseline with multi-task learning aided depth estimation module | $0.016 \pm 0.0075$ | $0.018 \pm 0.0090$ |
| Proposed complete method | $0.011 \pm 0.0050$ | $0.013 \pm 0.0060$ |

Compared with the baseline, the baseline with multi-task learning aided depth estimation module obtains 35.9% better estimation at average, showing that the multi-task learning framework can obviously improve the accuracy of pose estimation. It is known from the aforementioned comparison results of depth estimation, that the accuracy of depth estimation is improved to a certain extent via the multi-task learning framework. That is to say, the improvement of depth estimation accuracy can further promote the pose estimation by a large margin in the VO, indicating the significance of the multi-task learning framework for pose estimation.

As shown in Tab.3, the baseline with cycled GAN outperforms the baseline, due to the constraints and consistency from the cycle. In contrast to the limited accuracy gain in depth estimation, the preserved details are more important to image matching for pose regression, leading to a big improvement in pose estimation. In fact, this cycle has the same significance as a data augmentation approach because, at training time, the network learns to predict the depth map and pose not only from the images of the training in the forward half-cycle, but also from synthesized images in the backward half-cycle. Since the generator is used both to generate training data and to estimate depth and pose, the backward half-cycle prevents the forward half-cycle network from predicting inconsistent depth and pose, producing the cycled consistence to improve the estimation accuracy of the forward half-cycle network. Remarkably, the testing is performed using only the forward half-cycle network after the training has been finished. Therefore, the proposed baseline with cycled GAN does not increase the testing computation time but only the training complexity.

As illustrated in Tab.3, the baseline with robust cycled GAN can acquire 15.8% better pose estimation than baseline with cycled GAN at average, showing the effectiveness of brightness robust losses to pose estimation. Indeed, the proposed brightness robust losses, combining gradient-based loss and perceptual loss, can provide a regularization to the changes in illumination, decreasing essentially the weight of losses based on illumination differences between the synthesized image and target image. Thus, the brightness robust losses make the self-supervised signal more robust against illumination changes, resulting in high-precision pose estimation in foggy weather with local and global illumination changes.

In addition, the performance of the proposed complete method was evaluated compared with some state-of-the-art deep learning based monocular VO. The results of pose estimation is shown in Tab. 4 and Fig. 5, which show the proposed complete method can reach higher estimation accuracy than all the other approaches do in foggy weather.

Table 4. The pose estimation results by different methods

| Method | Absolute trajectory error on foggy KITTI sequence 09 (m) | Absolute trajectory error on foggy KITTI sequence 10 (m) |
| --- | --- | --- |
| SfMLearner [16] | 0.040±0.020 | 0.042±0.022 |
| Vid2Dpeth [30] | 0.029±0.016 | 0.030±0.018 |
| GeoNet [17] | 0.028±0.014 | 0.029±0.016 |
| Zhan et al. [13] | 0.027±0.012 | 0.028±0.014 |
| Proposed complete method | 0.011±0.0050 | 0.013±0.0060 |

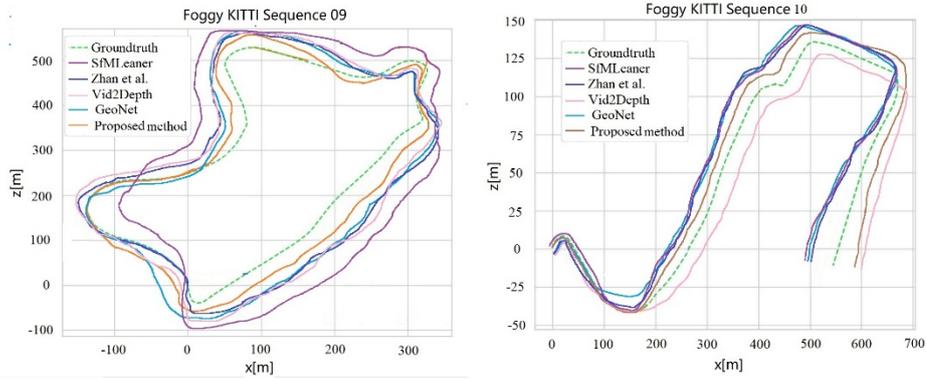

Figure 5. Sample trajectories by different methods

As shown in the above qualitative and quantitative comparisons, the proposed complete method acquires better pose estimation than other state-of-the-art methods due to the cycled GAN based architecture, robust losses against brightness changes, and multi-task learning framework, which demonstrates the high accuracy and strong robustness of the designed VO in foggy weather.

5 Conclusion

Self-supervised monocular VO has the potential to achieve high accuracy without ground truth in harsh environments due to the strong robustness in feature extraction. However, the positioning accuracy of most existing self-supervised monocular VO in foggy weather is too low to meet the practical requirements, because the adverse effect of fog on self-supervised loss and the ill posed problem of depth estimation have not been well solved.

To address these issues, this paper presents a high-precision self-supervised monocular VO, which is specifically designed for foggy weather. A cycled generative adversarial network is designed to obtain high-quality self-supervised loss via forcing the forward and backward half-cycle to output consistent estimation. In addition, gradient-based loss and perceptual loss are introduced to eliminate the interference of complex photometric change on self-supervised loss in foggy weather. A self-supervised multi-task learning aided depth estimation module is designed based on the strong correlation between the depth estimation and transmission map calculation of hazy images in foggy weather, aiming at solving the ill-posed problem of depth estimation, further leading to significantly improved pose estimation.

A synthetic foggy KITTI dataset was created based on the publicly-available dataset KITTI according to the method in [34]. The ablation experiments verified the effectiveness of the proposed cycled generative adversarial network, gradient-based loss, perceptual loss and self-supervised multi-task learning aided depth estimation module. The experimental results on the foggy KITTI dataset show that the proposed approach in this paper can obtain the best depth and pose estimation results compared with some representative existing monocular VO in the literature, indicating that the designed method is more suitable for navigation in foggy weather.

Acknowledgments

This work was supported by the Talent Award Scheme of Shanxi Province (Grant no. 304/18001618) and the Science Foundation of North University of China (Grant no.304/20170024).